\documentclass[runningheads]{llncs}
\usepackage{graphicx}

\usepackage{siunitx}
\usepackage{booktabs}
\usepackage{etoolbox}
\usepackage{cite}

\usepackage{xcolor}
\usepackage{hyperref}
\hypersetup{
    colorlinks=true,
    linkcolor=blue,
    filecolor=magenta,      
    urlcolor=cyan,
    citecolor=darkgray,
    breaklinks=true,
    pdftitle={The Enigma of Complexity},
    pdfpagemode=FullScreen,
}

\graphicspath{{Figures/}}

\begin{document}
\title{The Enigma of Complexity}
%
%
\author{Jon McCormack\inst{1}\orcidID{0000-0001-6328-5064} \and
Camilo Cruz Gambardella \inst{1}\orcidID{0000-0002-8245-6778} \and
Andy Lomas \inst{2}\orcidID{0000-0003-2177-3021}}
\authorrunning{J. McCormack et al.}
%
\institute{SensiLab, Monash University, Melbourne, Australia\\
\email{Jon.McCormack@monash.edu} \email{Camilo.CruzGambardella@monash.edu}\\
\url{https://sensilab.monash.edu} \and
Goldsmiths, University of London, London, UK\\
\email{andylomas@gmail.com}\\
\url{https://andylomas.com}}

\maketitle              
\begin{abstract}
In this paper we examine the concept of complexity as it applies to generative art and design. Complexity has many different, discipline specific definitions, such as complexity in physical systems (entropy), algorithmic measures of information complexity and the field of ``complex systems''. We apply a series of different complexity measures to three different generative art datasets and look at the correlations between complexity and individual aesthetic judgement by the artist (in the case of two datasets) or the physically measured complexity of 3D forms. Our results show that the degree of correlation is different for each set and measure, indicating that there is no overall ``better'' measure. However, specific measures do perform well on individual datasets, indicating that careful choice can increase the value of using such measures. We conclude by discussing the value of direct measures in generative and evolutionary art,  reinforcing recent findings from neuroimaging and psychology which suggest human aesthetic judgement is informed by many extrinsic factors beyond the measurable properties of the object being judged.

\keywords{Complexity \and aesthetic measure
\and generative art
\and generative design
\and evolutionary art
\and fitness measure.}
\end{abstract}
%
%
%
\section{Introduction}
\label{s:introduction}
\begin{quotation}
    ``The number of all the atoms that compose the world is immense but finite, and as such only capable of a finite (though also immense) number of permutations. In an infinite stretch of time, the number of possible permutations must be run through, and the universe has to repeat itself. Once again you will be born from a belly, once again your skeleton will grow, once again this same page will reach your identical hands, once again you will follow the course of all the hours of your life until that of your incredible death.''%
\flushright ---Jorge Luis Borges, \emph{The doctrine of cycles}, 1936
\end{quotation}

Complexity is a topic of endless fascination in both art and science. For hundreds of years scholars, philosophers and artists have sought to understand what it means for something to be ``complex'' and why we are drawn to complex phenomena and things. Today, we have many different understandings of complexity, from information theory, physics, and aesthetics \cite{Prigogine1980,Crutchfield1994b,GellMann1995,Wolfram2002}.

In this paper we again revisit the concept of complexity, with a view to understanding if it can be useful for the generative or evolutionary artist. The application of complexity measures and their relation to aesthetics in generative and evolutionary art are numerous (see e.g.~\cite{Johnson2019} for an overview). A number of researchers have tested complexity measures as candidates for fitness measures in evolutionary art systems for example. Here we are interested in the value of complexity to the individual artist or designer, not the system (or though that may benefit too). Put another way, we are asking what complexity can tell us about an individual artist's personal aesthetic taste or judgement, rather than the value of such measures in general. 

A long held intuition is that visual aesthetics are related to an artefact's order and complexity \cite{berlyne1971aesthetics, Klinger2000, machado2015computerized}.
From a human perspective, complexity is often regarded as the amount of ``processing effort'' required to make sense of an artefact. Too complex and the form becomes unreadable, too ordered and one quickly looses interest.
Birkhoff \cite{Birkhoff1933} famously formalised an aesthetic measure $M=O/C$, the ratio of order to complexity \cite{Birkhoff1933}, and similar approaches have built on this idea. To mention some examples, Berlyne and colleagues, defined visual complexity as ``irregularities in the spatial elements'' that compose a form \cite{papadimitriou2020spatial}, which lead to the formalisation of the relationship between pleasantness and complexity as an ``inverted-U'' \cite{berlyne1971aesthetics}. That is, by increasing the complexity of an artefact beyond the ``optimum'' value for aesthetic preference, its appeal starts to decline \cite{sun2014relationship}. Another example is Biderman's theory of ``geons'', which proposes that human understanding of spatial objects depends on how discernible its basic geometric components are \cite{Biederman1993, papadimitriou2020spatial} Thus, the harder an object is to decompose into primary elements, the more complex we perceive it is. This is the basis for some image compression techniques, which are also used as a measure of visual complexity \cite{Lakhal2020}. 

More recent surveys and analysis of computational aesthetics trace the history \cite{Greenfield2005, Hoenig2005} and current state of research in this area \cite{Johnson2019}. Other approaches introduce features such as symmetry as a counterbalance to complexity, situating aesthetic appeal somewhere within the range spanning between these two properties \cite{papadimitriou2020spatial}. The most recent approaches combine measures of algorithmic complexity with different forms of filtering or processing to eliminate noise but retain overall detail \cite{Zanette2018, Lakhal2020}.

Multiple attempts to craft automated methods for the aesthetic judgement of images have made use of complexity measures. Moreover, some of these show encouraging results. In this paper we test a selection of these methods on three different image datasets produced using generative art systems. All of the images in these datasets have their own ``aesthetic'' score as a basis for understanding the aesthetic judgements of the system's creator.   

\section{Complexity and Aesthetic Measure}

Computational methods used to calculate image complexity are based on the definitions of complexity described the previous section (Section \ref{s:introduction}): the amount of ``effort'' required to reproduce the contents of the image, as well as the way in which the patterns found in an image can be decomposed. Some methods have been proposed as useful measures of aesthetic appeal, or for predicting a viewer's preference for specific kinds of images.
In this section we outline the ones relevant for our research.

In the late 1990s Machado and Cardoso proposed a method to determine aesthetic value of images derived from their interpretation of the process that humans follow when experiencing an aesthetic artefact \cite{Machado98}. In their method the authors use a ratio of \emph{Image Complexity} -- a proxy for the complexity of the art itself -- to \emph{Processing Complexity} -- a proxy of the process humans use to make sense of an image --  as a representation of how humans perceive images. 

In 2010, den Heijer and Eiben compared four different aesthetic measures on a simple evolutionary art system \cite{den2010comparing}, including Machado and Cardoso's \emph{Image Complexity} / \emph{Processing Complexity} ratio, Ross \& Ralph's colour gradient bell curve, and the fractal dimension of the image. Their experiments demonstrated that, when used as fitness functions, different metrics yielded stylistically different results, indicating that each assessment method biases the particular image features or properties being evaluated. Interestingly, when interchanged -- when the results evolved with one metric are evaluated with another -- metrics showed different affinities, suggesting that regardless of the specificity of each individual measure, there are some commonalities between them.

\section{Experiments}
\label{s:experimets}
To try and answer our question about the role and value of complexity measures in developing generative or evolutionary art systems, we compared a variety of complexity measures on three different generative art datasets, evaluating them for correlation with human or physical measures of aesthetics and complexity.

\subsection{Complexity Measures}
\label{ss:complexity-measures}
We tested a number of different complexity measures described in the literature to see how they correlated with individual evaluations of aesthetics. We first briefly introduce each measure here and will go into more detail on specific measures later in the paper.

\begin{description}
    \item[Entropy ($S$):]{the image data entropy measured using the luminance histogram (base $e$).}
    \item[Energy ($E$):]{the data energy of the image.}
    \item[Contours ($T$):]{the  number  of  lines  required  to  describe component boundaries detected in the image. The image first undergoes a morphological binarisation (reduction to a binary image that differentiates component boundaries) before detecting the boundaries.}
    \item[Euler ($\gamma$):]{the morphological Euler number of the image (effectively a count of the number of connected regions minus the number of  holes). As with the $T$ measure, the image is first transformed using a morphological binarisation.}
    \item[Algorithmic Complexity ($C_a$):]{measure of the algorithmic complexity of the image using the method described in \cite{Lakhal2020}.} Effectively the compression ratio of the image using Lempil-Ziv-Welch lossless compression.
    \item[Structural Complexity ($C_s$):]{measure of the structural complexity, or ``noiseless entropy'' of an image using the method described in \cite{Lakhal2020}.}
    \item[Machardo-Cardoso Complexity ($C_{mc}$):]{a complexity measure used in \cite{machado2015}, without edge detection pre-processing.}
    \item[Machardo-Cardoso Complexity with edge detection ($C_{mc}^E$):]{the $C_{mc}$ measure with pre-processing of the image using a Sobel edge detection filter.}
    \item[Fractal Dimension ($D$):]{fractal dimension of the image calculated using the box-counting method \cite{forsythe2011predicting}.}
    \item[Fractal Aesthetic ($D_a$)]{aesthetic measure similar to that used in \cite{den_Heijer_2010}, based on the fractal dimension of the image fitted to a Gaussian curve with peak at 1.35. This value is chosen based on an empirical study of aesthetic preference for fractal dimension.}
\end{description}

While each of these measures is in some sense concerned with measuring image complexity, the basis of the measure for each is different. \emph{Entropy} ($S$) and \emph{Energy}($E$) measures are based on information theoretic understandings of complexity but concern only the distribution of intensity, while \emph{Contours} ($T$) and \emph{Euler} ($\gamma$) try to directly count the number of lines or features in the image, somewhat in line with perceptual notions of complexity. Lakhal et.~al's \emph{Algorithmic Complexity} ($C_a$) and \emph{Machardo \& Cardoso's Complexity} ($C_{mc}$) measures use algorithmic or Kolmogrov-like understandings of complexity, relying on image compression algorithms to proxy for visual complexity. Lakhal et.~al also define a \emph{Structural Complexity} measure ($C_s$) designed to address the limitations of algorithmic complexity measures in relation to high frequency noise or many fine details. This is achieved by a series of ``course-graining'' operations, effectively low-pass filtering the image to remove high frequency detail in both the spatial and intensity domains.  Finally, the fractal methods recognise self-similar features as proxies for complexity. They are based on past analysis of art images that demonstrated relationships between fractal dimension and aesthetics \cite{Peitgen1986,Taylor:1999,forsythe2011predicting}.

\subsection{Datasets}
\label{ss:datasets}

\begin{figure}
\begin{center}
  \begin{tabular}{ccc}
\includegraphics[width=0.3\textwidth]{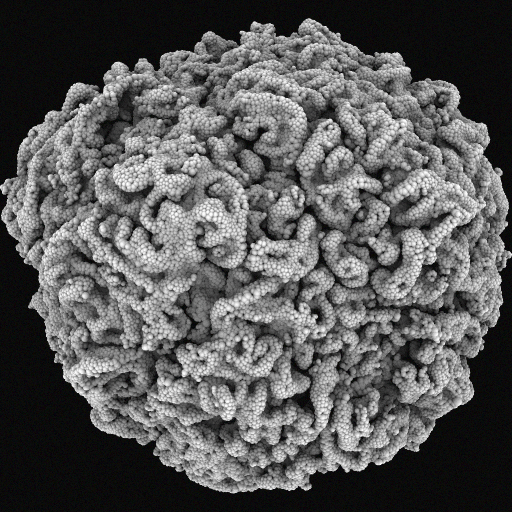} & \includegraphics[width=0.3\textwidth]{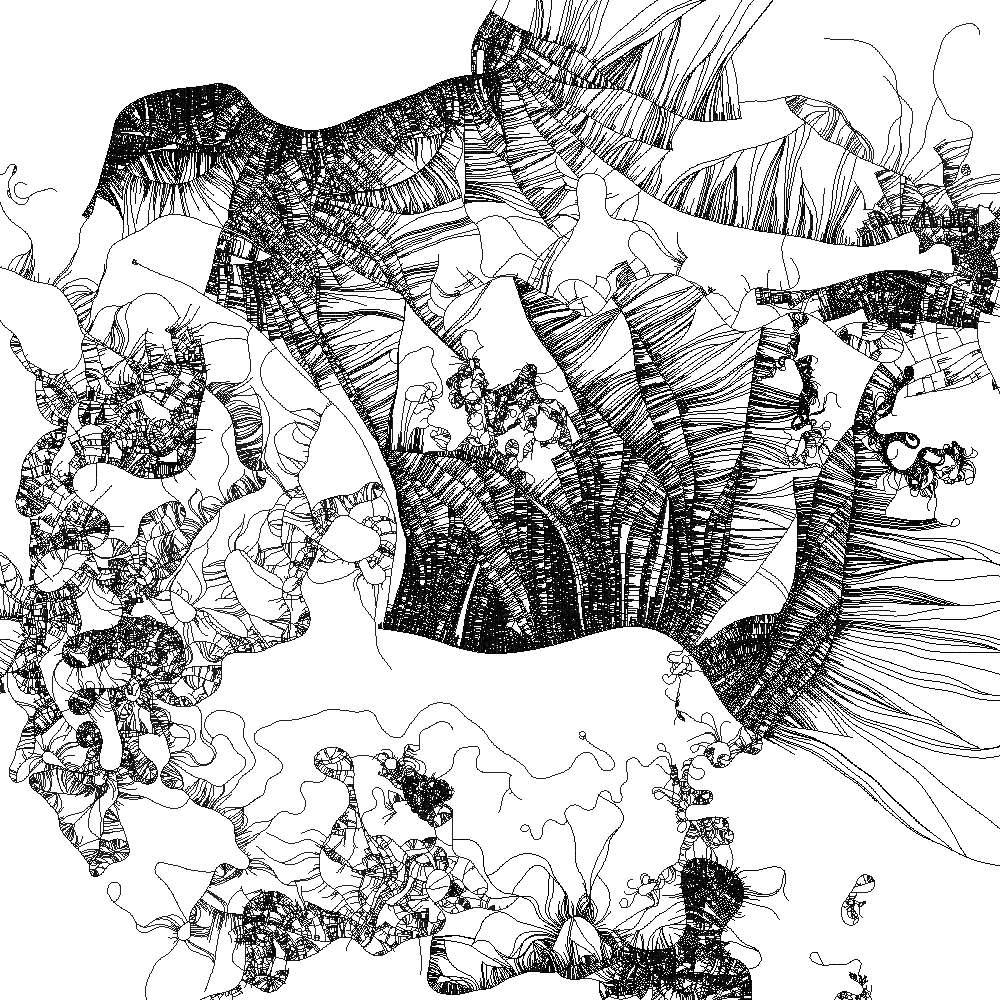} & \includegraphics[width=0.3\textwidth]{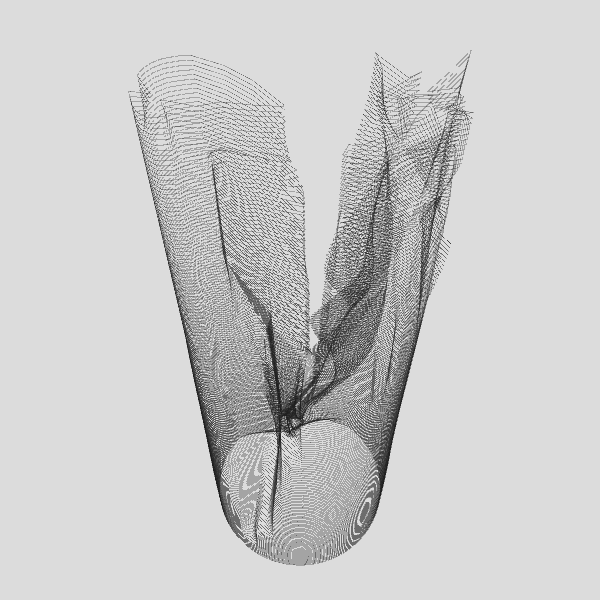} \\
a & b & c
  \end{tabular}
\end{center}
\caption{Example images from the Lomas (a), Line Drawing (b) and 3D DLA Forms (c) datasets.} \label{f:datasets}
\end{figure}

For the experiments described in this paper, we worked with three different generative art datasets (Figure \ref{f:datasets}). As the goal of this work was to understand the effectiveness of complexity measures in actual generative art applications, we wanted to work with artistic systems of demonstrated success, rather than invented or ``toy'' systems often used in this research. This allows us to understand the ecological validity \cite{Brunswik1956} of any system or technique developed. Ecological validity requires the assessment of creative systems in the typical environments and contexts under which they are actually developed and used, as opposed to laboratory or artificially constructed settings. It is considered an important methodology for validating research in the creative and performing arts \cite{Jausovec2011}. Additionally, all the datasets are open access, allowing others to validate new methods on the same data.

\subsubsection{Dataset 1: Andy Lomas' Morphogenetic Forms}
\label{sss:lomasDataSet}
This dataset \cite{LomasDS2020} consists of 1,774 images generated using a 3D morphogenetic form generation system, developed by computer artist Andy Lomas \cite{Lomas2016,Lomas2018}. Each image is a two-dimensional rendering ($512 \times 512$ pixels) of a three-dimensional form that has been algorithmically ``grown'' from 12 numeric parameters. The images were evolved using an \emph{Interactive Genetic Algorithm} (IGA)-like approach with the software \emph{Species Explorer} \cite{Lomas2016,Lomas2018}. As the 2D images, not the raw 3D models are evaluated by the artist, we perform our analysis similarly.

The dataset contains an integer numeric aesthetic rating score for each form (ranging from 0 to 10, with 1 the lowest and 10 the highest, 0 meaning a failure case where the generative system terminated without generating a form or the result was not rated). These ratings were all performed by Lomas, so represent his personal aesthetic preferences. Additionally, each form is categorised by Lomas into one of eight distinct categories (these were not used in the experiments described in this paper).

\subsubsection{Dataset 2: DLA 3D Prints}
\label{sss:dla-3d-prints}

\begin{figure}
\begin{centering}
\includegraphics[width=0.65\textwidth]{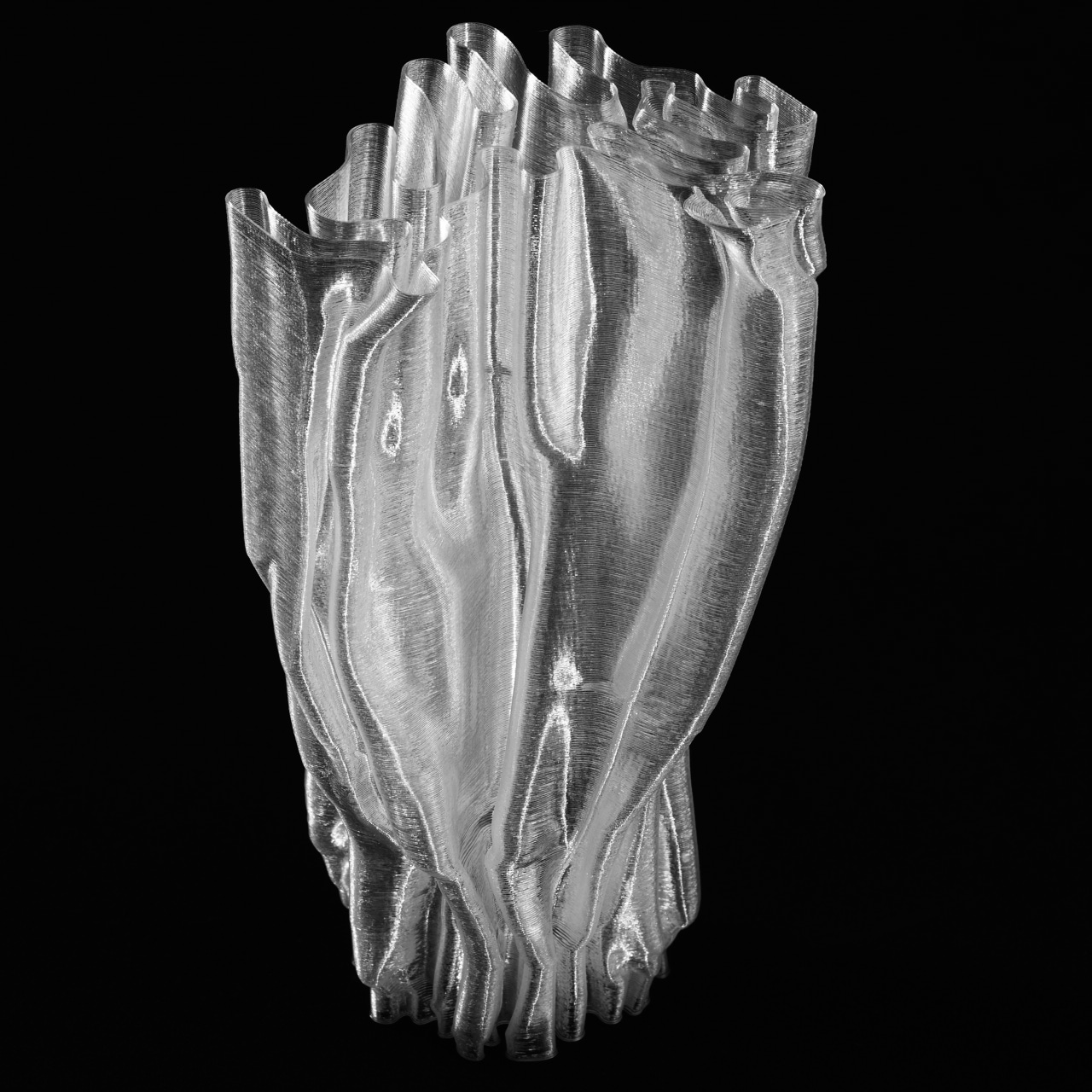}
\caption{Example 3D printed from from the DLA 3D Prints dataset.}
\label{f:heroPrint}
\end{centering}
\end{figure}

This dataset \cite{McCormack2021_DLADataset} consists of 2,500 3D forms created using a Differential Line Algorithm (DLA) based method \cite{barlow1989differential}. Multiple closed 2D line segments develop over time. At each time-step the geometry is captured and forms a sequential z-layer in a 3D form. After several hundred time-steps, the final 3D form is generated, suitable for 3D printing (Figure \ref{f:heroPrint}).  Each image is $600 \times 600$ pixels resolution. Images in this set are 3D line renderings of the final form, from a perspective projection and orthographic projection in the $xy$ plane. In the experiments described here we tested both the top-down orthographic images and perspective images, finding the perspective images gave better results and so are the ones reported here.

Rather than human-designated aesthetic measures, this dataset has a physically computed complexity measure. This measure is based on two geometric aspects of the 3D form:  \emph{convexity} (how much each layer deviates from its convex hull) and the quartile \emph{coefficient of dispersion} of angles between consecutive edges that make up each layer in the 3D form. These measures are calculated for each layer (weighted equally) and the final measure is the mean of all the layers in the form. This physical complexity measure appears to be a reasonable proxy to the visual complexity of the forms generated by the system.

\subsubsection{Dataset 3: Line Drawings}
\label{sss:line-drawings}
A set of 53 line drawings generated using an agent-based method based on the biological principles of niche construction \cite{McCormack2010,McCormackB09}. Each image is $1024 \times 1024$ pixels resolution. The dataset \cite{McCormack2021_ncDataset} also contains artist assigned aesthetic scores normalised to the range $[0,1]$.

\subsection{Settings and Measure Details}
\label{ss:settings}

Our preliminary investigations showed that some measures are sensitive to parameter settings. The structural complexity measure ($C_s$) has two parameters: $r_{cg}$, a course-grain filter radius (in pixels), $\delta \in [0,0.5]$ a threshold for determining the black to white pixel ratio, $\eta \in [0,1]$ (white if $\eta \le \delta$, grey if $\delta < \eta \le 1 - \delta$, black for $\eta > 1 - \delta$). In the original study, the authors \cite{Lakhal2020} used values $(r_{cg},\delta) = (7,0.23)$ for one set of test images (abstract textures generated by Fourier synthesis) and $(13,0.12)$ for the second set (abstract random boxes placed using an inverse of the fractal box counting method) for $256 \times 256$ resolution images. For the experiments described her we used $(r_{cg},\delta) = (5,0.23)$ as our image sizes were larger and the images contain significant high frequency detail.

\begin{figure}
\begin{center}
  \begin{tabular}{ccc}
\includegraphics[width=0.3\textwidth]{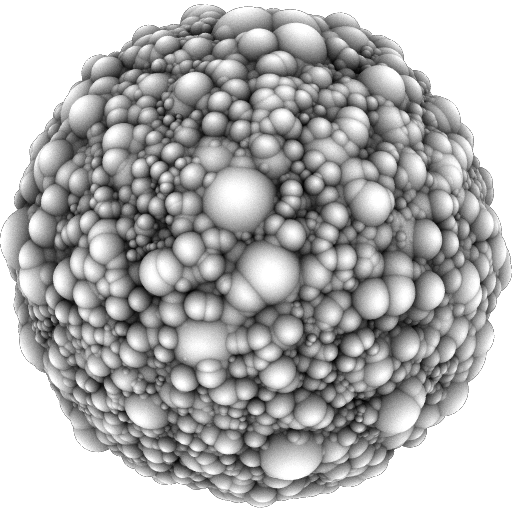} & \includegraphics[width=0.3\textwidth]{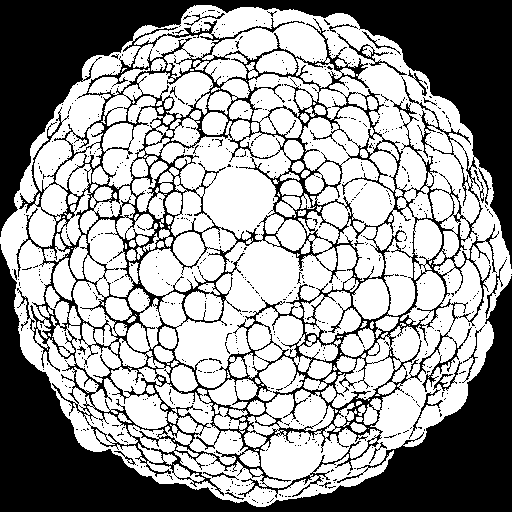} & \includegraphics[width=0.3\textwidth]{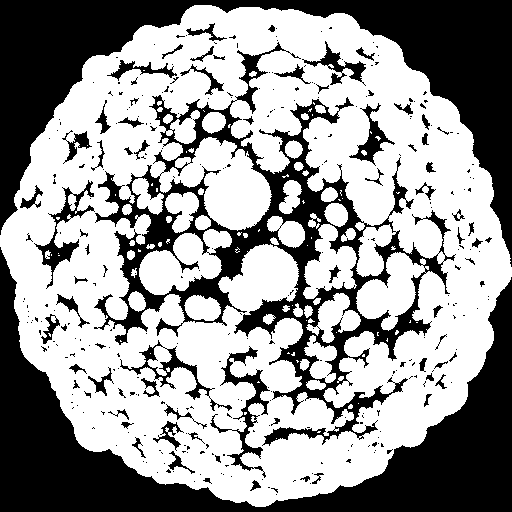} \\
Original Image & $r = 2$ & $r = 200$ \\
Fractal Dimension & 1.864 & 1.845
  \end{tabular}
\end{center}
\caption{The effect of different adaptive binarisation radii on an image from the Lomas dataset} \label{f:labin}
\end{figure}

For the fractal dimension measurements ($D,D_a$), images are pre-processed using a local adaptive binarisation process to convert the input image to a binary image (typically used to segment the foreground and background). A radius, $r$, is used to compute the local mean and standard deviation over $(2 r+1) \times (2 r+1)$ blocks centered on each pixel. Values above the mean of the $r$-range neighbourhood are replaced by 1, others by 0. Figure \ref{f:labin} shows a sample image from the Lomas dataset (left) with binary versions for $r = 2$ (middle) and $r = 200$ (right). Higher values of $r$ tend to reduce high frequency detail and result in a lower fractal dimension measurement. For the DLA 3D prints and Line Drawing datasets, which are already largely comprised of lines, the value of $r$ has negligible effect on the measurement.

Our Fractal Aesthetic Measure ($D_a$) is defined as:
\begin{equation}
    D_a(i) = exp(\frac{-(D(i) - p)^2}{2 \sigma^2}),
\end{equation}
where $p$ is the peek preference value for fractal dimension and $\sigma$ the width of the preference curve. $D_a$ returns a normalised aesthetic measure $\in [0,1]$. For the results reported here we used $(p,\sigma) = (1.35,0.2)$, based on prior findings for this preference \cite{Spehar2003}.

The Machardo-Cardoso Complexity measure ($C_{mc})$) is defined as:
\begin{equation}
C_{mc}(i) = RMS(i, f(i)) \times \frac{s(f(i))}{s(i)},
\end{equation}
where $i$ is the input image, $RMS$ a function that returns the root mean squared error between it's two arguments, $f$ a lossy encoding scheme for $i$ and $s$ a function that returns the size in bytes of its argument.\footnote{We adopted this measure as it specifically deals with complexity as defined in \cite{machado2015computerized}. Machardo \& Cardoso also define an aesthetic measure as the ratio of image complexity to processing complexity \cite{Machado98}, as used by den Heijer \& Eiben in their comparison of aesthetic measures \cite{den_Heijer_2010}.} For the lossy encoding scheme we used the standard JPEG image compression scheme with a compression level of 0.75 (0 is maximum compression).

\section{Results}
\label{s:results}
For each dataset we computed the full set of complexity measures (Section \ref{ss:complexity-measures}) on every image in the dataset, then computed the Pearson correlation coefficient between each measure and the human assigned aesthetic score (Lomas and Line Drawings datasets) or physically calculated complexity measure (DLA 3D Prints dataset).

\begin{table}
\caption{Lomas Datatset: Pearson's correlation coefficient values between image measurements and aesthetic score ($Sc$). The $C_{mc}$ complexity measure (bold) has the highest correlation with aesthetic score for this dataset. In all cases $p$-values are $< 1 \times 10^{-3}$}.
\label{tab:lomasCorrelations}
\robustify\bfseries
\begin{center}
\begin{tabular}{c|SSSSSSS[detect-weight]SSSS}
\toprule
 & {$S$} & {$E$} & {$T$} & {$\gamma$} & {$C_a$} & {$C_s$} & {$C_{mc}$} & {$C_{mc}^E$} & {$D$} & {$D_a$} & {Sc} \\
\midrule
$S$  &      1 &        &        &       &       &       &       &       &       &       & \\ 
$E$   & -0.989 &      1 &        &       &       &       &       &       &       &       & \\
$T$ &  0.425 & -0.375 &      1 &       &       &       &       &       &       &       & \\
$\gamma$    & -0.423 &  0.373 & -0.999 &     1 &       &       &       &       &       &       & \\
$C_a$    &  0.974 & -0.945 &  0.496 &-0.495 &     1 &       &       &       &       &       & \\
$C_s$    &  0.922 & -0.874 &  0.660 &-0.659 & 0.940 &     1 &       &       &       &       & \\
$C_{mc}$   &  0.793 & -0.732 &  0.590 &-0.589 & 0.907 & 0.860 &     1 &       &       &       & \\
$C_{mc}^E$ &  0.779 & -0.699 &  0.603 &-0.602 & 0.869 & 0.907 & 0.930 &     1 &       &       & \\
$D$     & -0.352 &  0.452 &  0.294 &-0.295 &-0.164 &-0.052 & 0.223 & 0.257 &    1  &       & \\
$D_a$    &  0.105 & -0.211 & -0.318 & 0.319 &-0.064 &-0.165 &-0.393 &-0.442 &-0.931 &     1 & \\
Sc    &  0.634 & -0.590 &  0.537 &-0.536 & 0.757 & 0.685 & \bfseries 0.873 & 0.774 & 0.284 & -0.389&     1 \\
\bottomrule
\end{tabular}
\end{center}
\end{table}

\begin{table}
\caption{DLA 3D Prints Datatset: Pearson's correlation coefficient values between image measurements and physically computed complexity score ($Sc$). The $C_{s}$ structural complexity measure (bold) has the highest correlation with aesthetic score for this dataset.}
\label{tab:formsCorrelations}
\robustify\bfseries
\begin{center}
\begin{tabular}{c|SSSSSS[detect-weight]SSSSS}
\toprule
 & {$S$} & {$E$} & {$T$} & {$\gamma$} & {$C_a$} & {$C_s$} & {$C_{mc}$} & {$C_{mc}^E$} & {$D$} & {$D_a$} & {Sc} \\
\midrule
$S$  &      1 &        &        &       &       &       &       &       &       &       & \\ 
$E$   & -0.995 &      1 &        &       &       &       &       &       &       &       & \\
$T$ &  0.857 & -0.880 &      1 &       &       &       &       &       &       &       & \\
$\gamma$    & 0.107 &  -0.083 & -0.363 &     1 &       &       &       &       &       &       & \\
$C_a$    &  0.953 & -0.956 &  0.936 &-0.106 &     1 &       &       &       &       &       & \\
$C_s$    &  0.882 & -0.892 &  0.925 &-0.204 & 0.942 &     1 &       &       &       &       & \\
$C_{mc}$   &  0.915 & -0.935 &  0.968 &-0.197 & 0.969 & 0.950 &     1 &       &       &       & \\
$C_{mc}^E$ &  0.914 & -0.935 &  0.961 &-0.188 & 0.965 & 0.954 & 0.999 &     1 &       &       & \\
$D$     & 0.928 &  -0.949 &  0.869 &0.012 &0.896 &0.801 & 0.898 & 0.895 &    1  &       & \\
$D_a$    &  -0.870 & -0.888 & -0.761 & -0.112 &-0.798 &-0.678 &-0.779 &-0.774 &-0.972 &     1 & \\
Sc    &  0.760 & -0.726 &  0.652 &-0.066 & 0.756 & \bfseries 0.774 & 0.704 & 0.706 & 0.550 & -0.434&     1 \\
\bottomrule
\end{tabular}
\end{center}
\end{table}

\begin{table}
\caption{Line Drawing Datatset: Pearson's correlation coefficient values between image measurements and aesthetic score ($Sc$). The Contours $T$ measure (bold) has the highest correlation with aesthetic score for this dataset.}
\label{tab:ncCorrelations}
\robustify\bfseries
\begin{center}
\begin{tabular}{c|SSS[detect-weight]SSSSSSSS}
\toprule
 & {$S$} & {$E$} & {$T$} & {$\gamma$} & {$C_a$} & {$C_s$} & {$C_{mc}$} & {$C_{mc}^E$} & {$D$} & {$D_a$} & {Sc} \\
\midrule
$S$  &      1 &        &        &       &       &       &       &       &       &       & \\ 
$E$   & -0.910 & 1        &       &       &       &       &       &       &       & \\
$T$ & 0.558 & -0.677 &      1 &       &       &       &       &       &       &       & \\
$\gamma$    & -0.559 &  0.677 & -1.000 &     1 &       &       &       &       &       &       & \\
$C_a$    &  0.994 & -0.934 &  0.541 &-0.541 &     1 &       &       &       &       &       & \\
$C_s$    &  0.576 & -0.717 &  0.474 &-0.474 & 0.618 &     1 &       &       &       &       & \\
$C_{mc}$   &  0.515 & -0.690 &  0.233 &-0.233 & 0.592 & 0.761 &     1 &       &       &       & \\
$C_{mc}^E$ &  0.648 & -0.811 &  0.312 &-0.312 & 0.712 & 0.822 & 0.927 &     1 &       &       & \\
$D$     & 0.580 &  -0.807 &  0.431 & -0.431 &0.640 & 0.835 & 0.867 & 0.914 &    1  &       & \\
$D_a$    &  -0.434 & 0.641 & -0.323 & 0.323 &-0.686 &-0.771 &-0.725 &-0.770 &-0.942 &     1 & \\
Sc    &  0.209 & -0.407 &  \bfseries 0.565 &-0.564 & 0.218 & 0.364 & 0.267 & 0.199 & 0.456 & -0.457 &     1 \\
\bottomrule
\end{tabular}
\end{center}
\end{table}

The results are shown for each dataset in Tables \ref{tab:lomasCorrelations} (Lomas), \ref{tab:formsCorrelations} (DLA 3D Prints) and \ref{tab:ncCorrelations} (Line Drawings) with the highest correlation measure shown in bold.

\begin{figure}
\begin{center}
  \begin{tabular}{cc}
\includegraphics[width=0.48\textwidth]{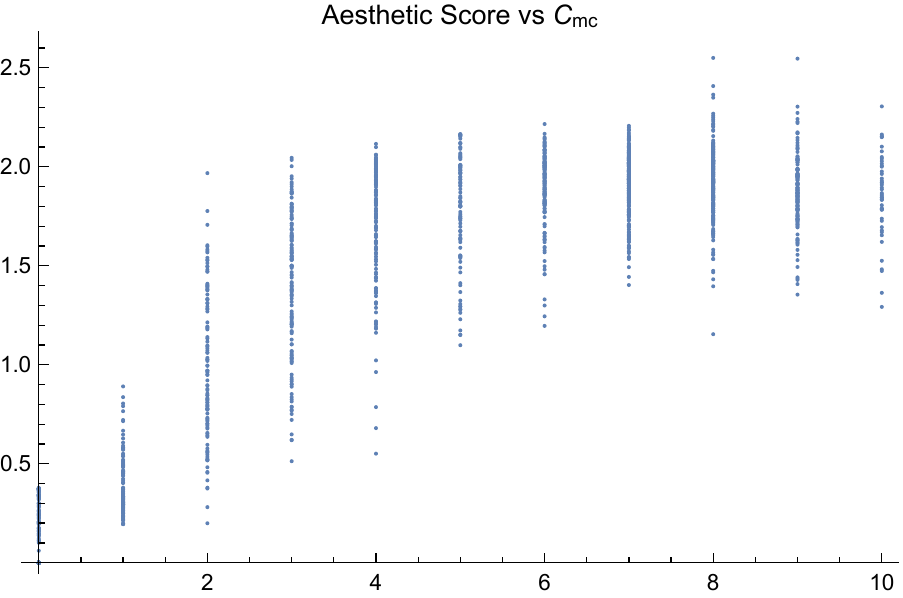} & \includegraphics[width=0.48\textwidth]{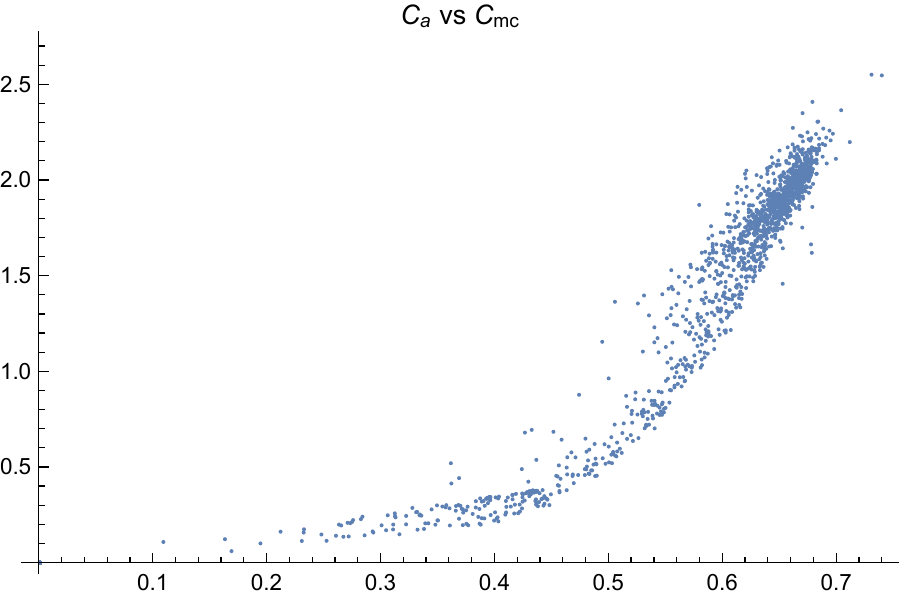}  \\
a & b
  \end{tabular}
\end{center}
\caption{Plots for the Lomas dataset showing the relationship between aesthetic score and $C_{mc}$ (a) and $C_a$ vs $C_{mc}$ (b). } \label{f:lomasPlots}
\end{figure}

As the tables show, a different complexity measure performed best for each dataset. For the \textbf{Lomas dataset} there is a strong correlation (0.873) between the artist assigned aesthetic score and the $C_{mc}$ complexity measure, and that all the algorithmic and structural complexity measures are highly correlated. This is to be expected since they all involve image compression ratios. It is further highlighted in Figure \ref{f:lomasPlots}, which shows a plot of aesthetic score vs $C_{mc}$ (a) and $C_s$ vs $C_{mc}$ (b). The banding in \ref{f:lomasPlots}a is due to the aesthetic scores being integers. A clear non-linear relationship between the complexity measures $C_s$ and $C_{mc}$ can be seen in \ref{f:lomasPlots}b.

Also of note is that fractal measures performed the worst of the measures tested. This seems to be confirmed visually: while certainly the images are complex (many are composed of 1 million or more cells) and have patterns at different scales, the patterns are not self-similar.

\begin{figure}
\begin{center}
  \begin{tabular}{cc}
\includegraphics[width=0.48\textwidth]{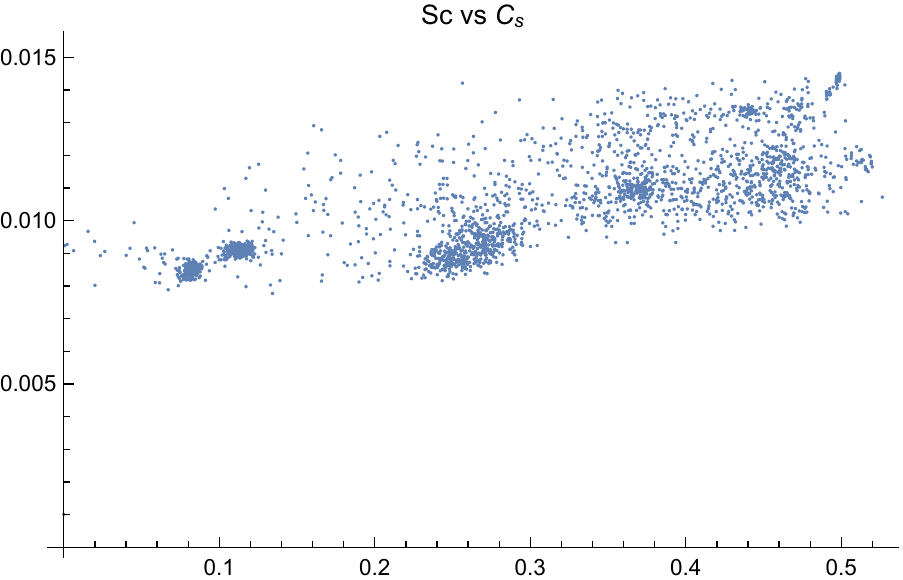} & \includegraphics[width=0.48\textwidth]{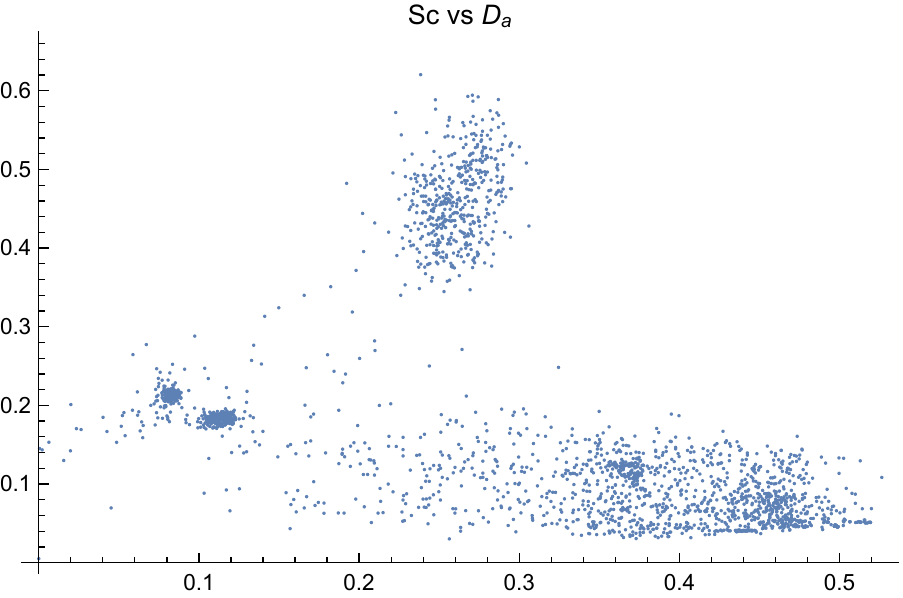}  \\
a & b
  \end{tabular}
\end{center}
\caption{Plots for the DLA 3D Prints dataset showing the relationship between physical complexity score ($Sc$) and $C_{s}$ (a) and $D_a$ (b). } \label{f:dlaPlots}
\end{figure}

For the \textbf{DLA 3D Prints} the most highly correlated measure was structural complexity ($C_s$) with a correlation of 0.774. The structural complexity aims to give a ``noiseless entropy'' measure by filtering high frequency spatial and intensity details. Given that the images are composed of many hundreds of thin lines stacked on top of each other, there is a significant amount of high frequency information, hence filtering is likely to give a better measure of real geometric details in each form. As can be seen in Figure \ref{f:dlaPlots} a clear correlation can be seen between the physical complexity ($Sc$) and Structural\footnote{Readers should not draw any direct relation between the terms ``structural'' and ``physical'' in relation to complexity used here. Structural refers to image structures, whereas physical refers to characteristics of the 3D form's line segments.} Complexity measure ($C_s$).
Again we note that the fractal measures ($D,D_a$) had the lowest correlation and that all the algorithmic complexity measures are highly correlated. As shown in Figure \ref{f:dlaPlots}b however, there appears to be a kind of bifurcation and clustering in the relationship between $Sc$ and $D_a$, indicating a more complex relationship between fractal dimension and complexity.

\begin{figure}
\begin{center}
  \begin{tabular}{c|c}
\includegraphics[width=0.48\textwidth]{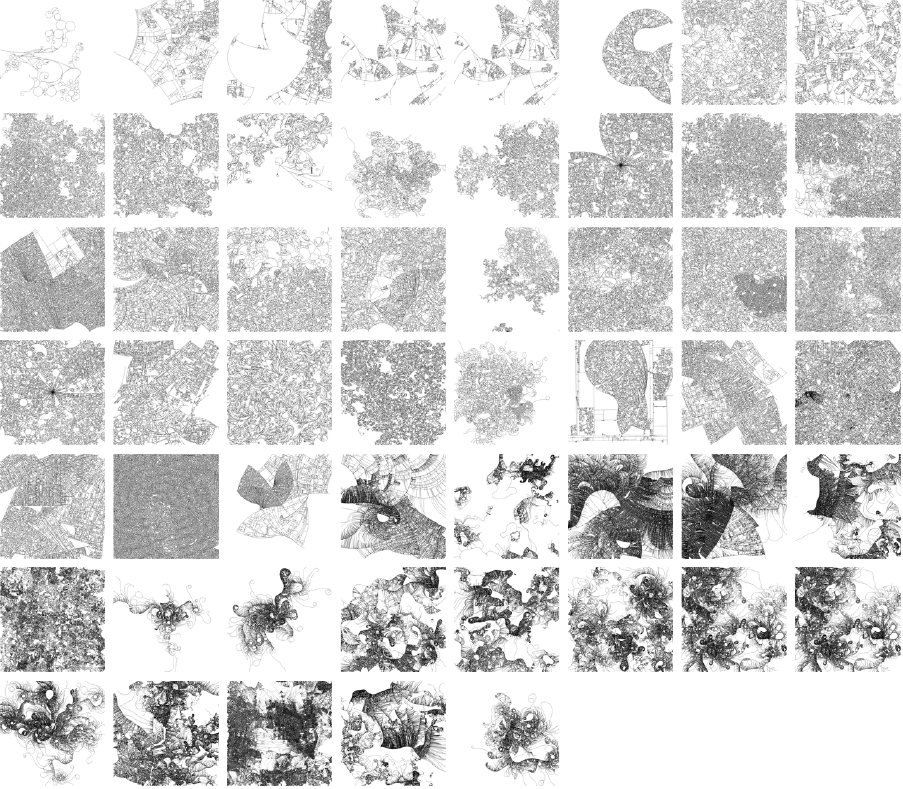} & \includegraphics[width=0.48\textwidth]{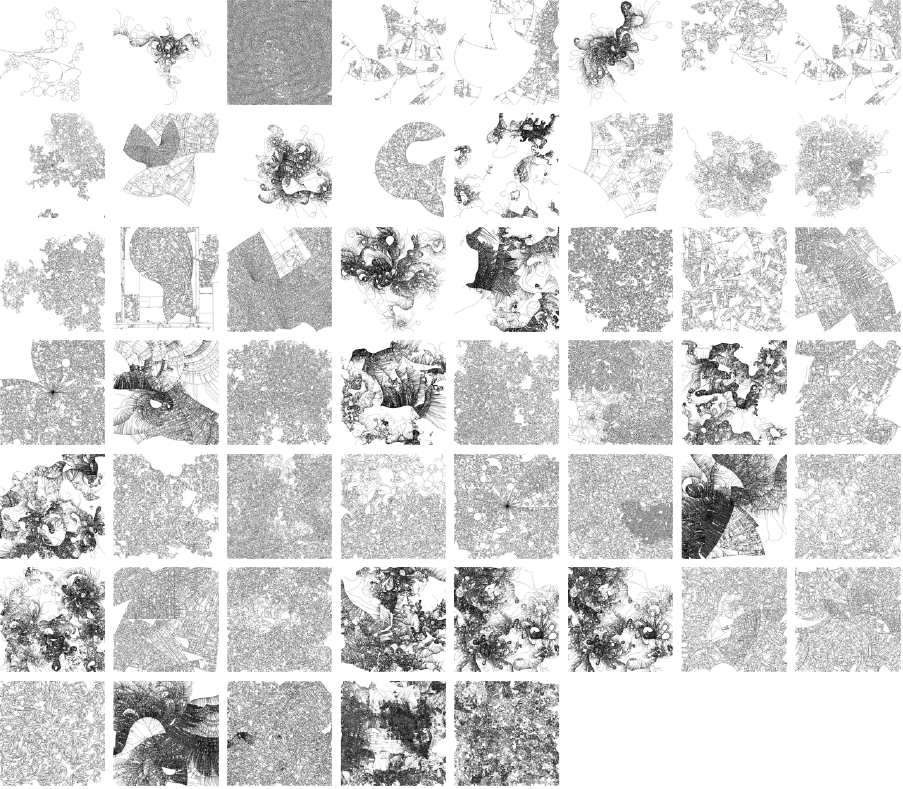}  \\
a & b
  \end{tabular}
\end{center}
\caption{Thumbnail grid of the entire Line Drawing dataset, ordered with increasing aesthetic score (lowest top left, highest bottom right) (a) and ordering by structural complexity ($C_s$). As the size of the dataset is relatively small in comparison with the others, the images can be shown in the figure.} \label{f:linePlots}
\end{figure}

The \textbf{Line Drawing dataset} exhibited quite different results over the previous two. Here the Contours ($T$) measure had the highest, but only moderate, correlation with artist-assigned aesthetic scores (0.565). Given the nature of the drawings, measures designed to capture morphological structure seem most appropriate for this dataset. It is also interesting to note that the algorithmic complexity measures perform relatively poorly in this case. The original basis for the drawings came from the use of niche construction as a way to generate density variation in the images. The dataset contains images both with and without the use of niche construction, and generally those with niche construction are more highly ranked than those without. Figure \ref{f:linePlots} shows the entire dataset ordered in terms of artist-assigned aesthetic score (a) and structural complexity (b). The drawings with niche construction are easy to see as they are more highly ranked than those without. The structural complexity measure has greater difficulty in differentiating them (b).

With this in mind, we ran an additional image measure on this dataset that looks at asymmetry in intensity distribution (\emph{Skew}). Since the niche construction process results in contrasting areas of high and low density it was hypothesised that this measure might be able to better capture the differences. This measure had a correlation of 0.583 ($p = 4.5 \times 10^{-6})$, so better than any of the other measures, but still only mildly correlated.

\section{Discussion}
\label{s:discussion}

Our results show that there appears to be no single measure that is best to quantify image complexity in the the context of generative art. Hence it seems wise to select a measure most appropriate to the style or class of imagery or form being generated. 

It is also important to point out that, in general, computer synthesised imagery and in particular images generated by algorithmic methods, have important characteristics that differ from other images, such as photographs or paintings. Apart from any semantic differences or differentiation between figurative and abstract, intensity and spatial distributions in computer synthesised images differ from real world images. This is one reason why we selected datasets that are specific to the application of these measures (generative art and design), rather than human art datasets in general, for example.

The rational for this research was to further the question: \emph{how can complexity measures be usefully employed in generative and evolutionary art and design?} Based on the results presented in this paper, our answer is that -- if chosen appropriately -- they can be valuable aids in  course-level discrimination. Additionally, they are quite quick to compute and work without prior training or exposure to large numbers of examples or training sets. So, for example, they could be helpful in filtering or ranking individuals in an IGA or used to help classify or select individuals for further enhancement. However they are insufficient as fully autonomous fitness measures -- the human designer remains a vital and fundamental part of any aesthetic evaluation.

\subsection{Aesthetic Judgement}
\label{ss:aestheticJudgement}

In Section \ref{s:introduction} we discussed possible relationships between complexity measures and aesthetics. It is worth reflecting further here on this relationship and the long-held ``open problem'' for evolutionary and generative art of quantifying aesthetic fitness \cite{McCormack2005a}.

In the last decade or so, the biggest advances in the understanding of computational and human aesthetic judgements have come from (i) large, open access datasets of imagery with associated human aesthetic rankings and (ii) psychological and neuroscience discoveries on the mechanisms of forming an aesthetic judgement and what constitutes aesthetic experience.

In a recent paper, Skov summarised aesthetic appreciation from the perspective of neuroimaging \cite{Skov2019}. Some of the key findings included neuroscientific evidence suggesting that ``aesthetic appreciation is not a distinct neurobiological process assessing certain objects, but a general system, centered on the mesolimbic reward circuit, for assessing the hedonic value of any sensory object'' \cite{Skov2019}. Another important finding was that hedonic values are not solely determined by object properties. They are subject to numerous extrinsic factors outside the object itself. Similar claims have come from psychological models \cite{Leder2014}. These findings suggest that any algorithmic measure of aesthetics which only considers an object's visual appearance ignores many other extrinsic factors that humans use to form an aesthetic judgement. Hence they are unlikely to correlate strongly with human judgements.

Our results appear to tally with these findings. Complexity measures, carefully chosen for specific styles or types of generative art can capture some broad aspects of personal aesthetic judgement, but they are insufficient alone to fully replace human judgement and discretion. Using other techniques, such as deep learning, \emph{may} result in slightly better correlation to individual human judgement \cite{McCormackLomas2020b}, however such systems require training on large datsets which can be tedious and time-consuming for the artist and still do not do as well as the trained artist's eye in resolving aesthetic decisions.

\section{Conclusion}
\label{s:conclusion}
Making and appreciating art is a shared human experience. Computers can expand and grow the creative possibilities available to artists and audiences. The fact that humans artists are successfully able to create and communicate artefacts of shared aesthetic value indicates some shared concept of this value between people and cultures. Could machines ever share such concepts? This remains an open question, but evidence suggests that achieving such a unity would require consideration of factors beyond the quantifiable properties of objects themselves.

In this paper we have examined the relationship between complexity measures and personal or specific understandings of aesthetics. Our results suggest that some measures can serve as crude proxies for personal visual aesthetic judgement but the measure itself needs to be carefully selected. Complexity remains an enigmatic and contested player in the long-term game of computational aesthetics.

%
%
%
\bibliographystyle{splncs04}
\bibliography{fullreferences.bib}

\begin{thebibliography}{10}
\providecommand{\url}[1]{\texttt{#1}}
\providecommand{\urlprefix}{URL }
\providecommand{\doi}[1]{https://doi.org/#1}

\bibitem{barlow1989differential}
Barlow, P., Brain, P., Adam, J.: Differential growth and plant tropisms: a
  study assisted by computer simulation. In: Differential Growth in Plants, pp.
  71--83. Elsevier (1989)

\bibitem{berlyne1971aesthetics}
Berlyne, D.E.: Aesthetics and psychobiology. Appleton-Century-Crofts, New York
  (1971)

\bibitem{Biederman1993}
Biederman, I.: Geon theory as an account of shape recognition in mind and
  brain. The Irish Journal of Psychology  \textbf{14}(3),  314--327 (1993)

\bibitem{Birkhoff1933}
Birkhoff, G.D.: Aesthetic Measure. Harvard University Press, Cambridge, MA
  (1933)

\bibitem{Brunswik1956}
Brunswik, E.: Perception and the representative design of psychological
  experiments. University of California Press, Berkley and Los Angeles, CA, 2nd
  edn. (1956)

\bibitem{Crutchfield1994b}
Crutchfield, J.P.: Complexity: Metaphors, Models, and Reality, vol.~XIX, chap.
  Is Anything Ever New?: Considering Emergence, pp. 479--497. Addison-Wesley,
  Redwood City (1994)

\bibitem{forsythe2011predicting}
{Forsythe}, A., {Nadal}, M., {Sheehy}, N., {Cela-Conde}, C.J., {Sawey}, M.:
  Predicting beauty: fractal dimension and visual complexity in art. British
  Journal of Psychology  \textbf{102}(1),  49--70 (2011)

\bibitem{GellMann1995}
Gell-Mann, M.: What is complexity? Complexity  \textbf{1}(1),  16--19 (1995)

\bibitem{Greenfield2005}
Greenfield, G.: {On the Origins of the Term Computational Aesthetics}. In:
  Neumann, L., Sbert, M., Gooch, B., Purgathofer, W. (eds.) Computational
  Aesthetics in Graphics, Visualization and Imaging. pp. 9--12. The
  Eurographics Association (2005). \doi{10.2312/COMPAESTH/COMPAESTH05/009-012}

\bibitem{den_Heijer_2010}
den Heijer, E., Eiben, A.E.: Comparing aesthetic measures for evolutionary art.
  In: Applications of Evolutionary Computation, pp. 311--320. Springer Berlin
  Heidelberg (2010). \doi{10.1007/978-3-642-12242-2\_32},
  \url{https://doi.org/10.1007/2F978-3-642-12242-2\_32}

\bibitem{den2010comparing}
den Heijer, E., Eiben, A.E.: Comparing aesthetic measures for evolutionary art.
  In: European Conference on the Applications of Evolutionary Computation. pp.
  311--320. Springer, Berlin, Heidelberg (2010)

\bibitem{Hoenig2005}
Hoenig, F.: {Defining Computational Aesthetics}. In: Neumann, L., Sbert, M.,
  Gooch, B., Purgathofer, W. (eds.) Computational Aesthetics in Graphics,
  Visualization and Imaging. The Eurographics Association (2005).
  \doi{10.2312/COMPAESTH/COMPAESTH05/013-018}

\bibitem{Jausovec2011}
Jausovec, N., Jausovec, K.: Brain, creativity and education. The Open
  Educational Journal  \textbf{4},  50--57 (2011)

\bibitem{Johnson2019}
Johnson, C.G., McCormack, J., Santos, I., Romero, J.: Understanding aesthetics
  and fitness measures in evolutionary art systems. Complexity
  \textbf{2019}(Article ID 3495962),  14 pages (2019).
  \doi{https://doi.org/10.1155/2019/3495962},
  \url{https://doi.org/10.1155/2019/3495962}

\bibitem{Klinger2000}
Klinger, A., Salingaros, N.A.: A pattern measure. Environment and Planning B:
  Planning and Design  \textbf{27}(4),  537--547 (2000)

\bibitem{Lakhal2020}
Lakhal, S., Darmon, A., Bouchaud, J.P., Benzaquen, M.: Beauty and structural
  complexity. Phys. Rev. Research  \textbf{2},  022058 (Jun 2020).
  \doi{10.1103/PhysRevResearch.2.022058},
  \url{https://link.aps.org/doi/10.1103/PhysRevResearch.2.022058}

\bibitem{Leder2014}
Leder, H., Nadal, M.: Ten years of a model of aesthetic appreciation and
  aesthetic judgments: The aesthetic episode -- developments and challenges in
  empirical aesthetics. British Journal of Psychology  \textbf{105},  443--464
  (2014)

\bibitem{Lomas2016}
Lomas, A.: Species explorer: An interface for artistic exploration of
  multi-dimensional parameter spaces. In: Bowen, J., Lambert, N., Diprose, G.
  (eds.) Electronic Visualisation and the Arts (EVA 2016). Electronic Workshops
  in Computing (eWiC), BCS Learning and Development Ltd., London (12th-14th
  July 2016)

\bibitem{Lomas2018}
Lomas, A.: On hybrid creativity. Arts  \textbf{7}(3), ~25 (2018).
  \doi{https://doi.org/10.3390/arts7030025}

\bibitem{Machado98}
Machado, P., Cardoso, A.: Computing aesthetics. In: Proceedings of the
  Brazilian Symposium on Artificial Intelligence, SBIA-98. pp. 219--229.
  Springer-Verlag (1998)

\bibitem{machado2015}
Machado, P., Romero, J., Nadal, M., Santos, A., Correia, J., Carballa, A.:
  Computerized measures of visual complexity. Acta Psychologica  \textbf{160},
  43--57 (September 2015). \doi{https://doi.org/10.1016/j.actpsy.2015.06.005}

\bibitem{machado2015computerized}
Machado, P., Romero, J., Nadal, M., Santos, A., Correia, J., Carballal, A.:
  Computerized measures of visual complexity. Acta psychologica  \textbf{160},
  43--57 (2015)

\bibitem{McCormack2005a}
McCormack, J.: Open problems in evolutionary music and art. In: Rothlauf, F.,
  Branke, J., Cagnoni, S., Corne, D.W., Drechsler, R., Jin, Y., Machado, P.,
  Marchiori, E., Romero, J., Smith, G.D., Squillero, G. (eds.) EvoWorkshops.
  Lecture Notes in Computer Science, vol.~3449, pp. 428--436. Springer (2005)

\bibitem{McCormack2010}
McCormack, J.: Enhancing creativity with niche construction. In: Fellerman, H.,
  D{\"o}rr, M., Hanczyc, M.M., Laursen, L.L., Maurer, S., Merkle, D., Monnard,
  P.A., Stoy, K., Rasmussen, S. (eds.) Artificial Life XII. pp. 525--532. MIT
  Press, Cambridge, MA (2010)

\bibitem{McCormack2021_ncDataset}
McCormack, J.: {Niche Constructions Generative Art Dataset} (1 2021),
  \url{https://bridges.monash.edu/articles/dataset/Niche_Constructions_Generative_Art_Dataset/13662383}

\bibitem{McCormackB09}
McCormack, J., Bown, O.: Life's what you make: Niche construction and
  evolutionary art. In: Giacobini, M., Brabazon, A., Cagnoni, S., Caro, G.A.D.,
  Ek{\'a}rt, A., Esparcia-Alc{\'a}zar, A., Farooq, M., Fink, A., Machado, P.,
  McCormack, J., O'Neill, M., Neri, F., Preuss, M., Rothlauf, F., Tarantino,
  E., Yang, S. (eds.) EvoWorkshops. Lecture Notes in Computer Science,
  vol.~5484, pp. 528--537. Springer (2009)

\bibitem{McCormack2021_DLADataset}
McCormack, J., Gambardella, C.C.: {DLA Form Generation dataset} (1 2021).
  \doi{10.26180/13663400.v1},
  \url{https://bridges.monash.edu/articles/dataset/DLA_Form_Generation_dataset/13663400}

\bibitem{LomasDS2020}
McCormack, J., Lomas, A.: Andy lomas generative art dataset.
  \doi{10.5281/zenodo.4047222}, \url{https://doi.org/10.5281/zenodo.4047222}

\bibitem{McCormackLomas2020b}
McCormack, J., Lomas, A.: Deep learning of individual aesthetics. Neural
  Computing and Applications  \textbf{33}(1),  3--17 (2020).
  \doi{https://doi.org/10.1007/s00521-020-05376-7}

\bibitem{papadimitriou2020spatial}
Papadimitriou, F.: Spatial complexity, visual complexity and aesthetics. In:
  Spatial Complexity, pp. 243--261. Springer (2020)

\bibitem{Peitgen1986}
Peitgen, H.O., Richter, P.H.: The beauty of fractals: images of complex
  dynamical systems. Springer-Verlag, Berlin; New York (1986)

\bibitem{Prigogine1980}
Prigogine, I.: From being to becoming: time and complexity in the physical
  sciences. W. H. Freeman, New York (1980)

\bibitem{Skov2019}
Skov, M.: Aesthetic appreciation: The view from neuroimaging. Empirical Studies
  of the Arts  \textbf{37}(2),  220--248 (2019).
  \doi{10.1177/0276237419839257},
  \url{https://doi.org/10.1177/0276237419839257}

\bibitem{Spehar2003}
Spehar, B., Clifford, C.W.G., Newell, B.R., Taylor, R.P.: Universal aesthetic
  of fractals. Computers \& Graphics  \textbf{27}(5),  813--820 (2003)

\bibitem{sun2014relationship}
Sun, L., Yamasaki, T., Aizawa, K.: Relationship between visual complexity and
  aesthetics: application to beauty prediction of photos. In: Agapito, L.,
  Bronstein, M., Rother, C. (eds.) Computer Vision - ECCV 2014 Workshops. ECCV
  2014. Lecture Notes in Computer Science. pp. 20--34. Springer, Cham (2014)

\bibitem{Taylor:1999}
Taylor, R.P., Micolich, A.P., Jonas, D.: Fractal analysis of {P}ollock's drip
  paintings. Nature  \textbf{399}, ~422 (1999)

\bibitem{Wolfram2002}
Wolfram, S.: A new kind of science. Wolfram Media, Champaign, IL (2002)

\bibitem{Zanette2018}
Zanette, D.H.: Quantifying the complexity of black-and-white images. PLoS ONE
  \textbf{13}(11),  e0207879 (2018).
  \doi{https://doi.org/10.1371/journal.pone.0207879}

\end{thebibliography}
\end{document}